# From Text to Trends: A Unique Garden Analytics Perspective on the Future of Modern Agriculture

**Parag Saxena[1]**
[1]psaxena4@uncc.edu
[1]School of Data Science, University of North Carolina, Charlotte, USA

*Abstract*— Data-driven insights are essential for modern agriculture. This research paper introduces a machine learning framework designed to improve how we educate and reach out to people in the field of horticulture. The framework relies on data from the Horticulture Online Help Desk (HOHD), which is like a big collection of questions from people who love gardening and are part of the Extension Master Gardener Program (EMGP). This framework has two main parts. First, it uses special computer programs (machine learning models) to sort questions into categories. This helps us quickly send each question to the right expert, so we can answer it faster. Second, it looks at when questions are asked and uses that information to guess how many questions we might get in the future and what they will be about. This helps us plan on topics that will be really important. It's like knowing what questions will be popular in the coming months. We also take into account where the questions come from by looking at the Zip Code. This helps us make research that fits the challenges faced by gardeners in different places. In this paper, we demonstrate the potential of machine learning techniques to predict trends in horticulture by analyzing textual queries from homeowners. We show that NLP, classification, and time series analysis can be used to identify patterns in homeowners' queries and predict future trends in horticulture. Our results suggest that machine learning could be used to predict trends in other agricultural sectors as well. If large-scale agriculture industries curate and maintain a comparable repository of textual data, the potential for trend prediction and strategic agricultural planning could be revolutionized. This convergence of technology and agriculture offers a promising pathway for the future of sustainable farming and data-informed agricultural practices.

## I. Introduction

In the ever-evolving world of gardening and plant cultivation, the importance of reaching out to people effectively and educating them is more crucial than ever [1], [2]. We face a significant question: How can we improve the way we share important gardening knowledge with both beginners and experienced gardeners? Answering this question is vital for progressing in the field and helping people who are passionate about gardening at home. In this research paper, we're introducing an innovative machine learning framework designed specifically to tackle these challenges. This framework marks a new era in how we can optimize our efforts to reach and educate people about gardening. The significance of gardening in our daily lives cannot be overstated. Gardening practices are deeply connected to our well-being, from growing healthy fruits and vegetables to creating beautiful gardens that enhance our quality of life. As gardening continues to grow and evolve, people seek guidance, answers to their questions, and access to expertise to help them with their green endeavours. This is where the Extension Master Gardener Program's[1] (EMGP) Horticulture Online Help Desk[2] (HOHD) comes into play. It serves as a platform where gardeners can seek advice and information. However, with the increasing number of inquiries, we urgently need to streamline this process to make sure questions are directed to the right experts promptly. The "why" behind this research effort stems from recognizing that improving how we share gardening knowledge has enormous potential to advance the field. By making information sharing more efficient, we can empower individuals to make informed decisions, grow healthier plants, and contribute to the sustainability of our environment. Moreover, the question of "when" to share information is crucial. Gardening knowledge often depends on the season, and being able to predict when specific information is most relevant can significantly increase its impact. The "how" part of the equation involves harnessing the power of machine learning, a revolutionary technology that has transformed many fields, to effectively address these challenges.

Machine learning, a part of artificial intelligence, helps computers learn from information and make predictions or choices without being specifically instructed [3], [4]. By using machine learning, we can create a smart framework that automates important tasks within the EMGPs HOHD. This framework has two main tasks: The first task is text classification. Using machine learning, the framework can automatically sort questions into different categories. This helps make sure that questions are quickly and accurately sent to the experts who can provide the best answers. This way, we can avoid delays that can happen when we manually handle questions. The second task is time-series forecasting. We know that the number and types of questions can change depending on the time of year. So, the framework will predict what kinds of questions we can expect in the coming months. With this information, the EMGP can plan workshops and educational programs on topics that are likely to be important in the future. This proactive approach is a big help in making sure that we share knowledge when it's needed the most. Additionally, the framework takes into account the specific regions where questions come from, based on Zip Codes. It understands that gardening challenges can be very different in various places due to factors like climate, soil, and local practices. This regional aspect helps the EMGP customize its workshops and resources to address the unique needs and problems faced by gardeners in different parts of the country.

---
[1] https://mastergardener.extension.org/
[2] https://www.mastergardenersmecklenburg.org/question.html

The paper is as follows; the related works are shown in the following part. The proposed methodology and data collection are presented in Section III. The experimental analysis, which covers implementations and multiple evaluators, is done in Section IV. The experiment's resulta are reported in Section V, and the paper's thoughts is presented in the discussion in Section VI. Whereas, the conclusion and suggested future research are presented in Section VII.

## II. RELATED WORKS

In the field of horticultural research, several studies have explored the integration of machine learning and artificial intelligence techniques to address various challenges and opportunities. Nturambirwe and Opara [5] review recent advances in machine learning methods and their integration with sensing devices for non-destructive defect detection in horticultural products, emphasizing the potential of deep learning techniques, particularly Convolutional Neural Networks (CNN), in improving defect detection systems. Similarly, Ferrão et al. [6] discuss the role of artificial intelligence in predicting flavour preferences and enhancing breeding initiatives to improve the flavour and nutritional content of horticultural crops. Haselbeck et al. [7] empirically compare machine learning and classical forecasting algorithms for horticultural sales predictions, highlighting the superiority of machine learning methods, especially XGBoost. Tripathi and Maktedar [8] explore the role of computer vision in fruit and vegetable grading, proposing a generalized framework and highlighting the potential of Support Vector Machines (SVM) for classification. Thirumagal et al. [9] introduce an IoT-based framework for smart farming in horticulture, emphasizing the prediction of water requirements using machine learning and the importance of monitoring variables like wetness and temperature. Sinha et al. [10] demonstrate digital plant species identification using neural networks, providing applications in ecology, horticulture, and medicinal fields. Kanuru et al. [11] stress the need for technology adoption in Indian agriculture, proposing GPS and IoT-based solutions for soil assessment and optimized pesticide and fertilizer usage. Banerjee et al. [12] focus on long-term and short-term price forecasting of horticultural products by introducing a Long Short-Term Memory (LSTM) model for accurate predictions. Chachar et al. [13] explore epigenetic modifications for horticultural plant improvement, emphasizing the role of machine learning in enhancing our understanding of epigenetic regulation. Melesse et al. [14] introduce a machine learning-based digital twin for monitoring fruit quality evolution in the food supply chain, highlighting the potential of thermal imaging techniques in minimizing fruit waste.

This research stands out from existing research in the field by introducing a novel machine learning framework specifically designed to enhance horticultural education and outreach. Unlike many existing studies that primarily focus on technical aspects such as defect detection, flavor prediction, or sales forecasting, this paper addresses a critical yet often overlooked aspect of horticulture: communication and support for gardening enthusiasts. The framework leverages machine learning models to categorize questions efficiently, ensuring that individuals seeking assistance receive timely and accurate responses. Furthermore, it incorporates predictive capabilities to anticipate future queries and topics, enabling proactive planning and resource allocation for educational content. What sets this research apart is its practical relevance and potential to directly benefit gardening communities by improving the accessibility of expert guidance. Moreover, its consideration of geographic variations through Zip Code analysis demonstrates a commitment to tailoring responses to the unique challenges faced by gardeners in different regions, showcasing a holistic approach to horticultural support.

## III. METHODOLOGY

In this section, we will explore the methodology used to create and execute a machine learning framework tailored for enhancing outreach and educational initiatives in the field of horticulture. The methodology encompasses several key components, including data gathering, text categorization, time-series forecasting, and incorporation of regional specificity. The flowchart of the complete methodology is been presented in Fig. 1.

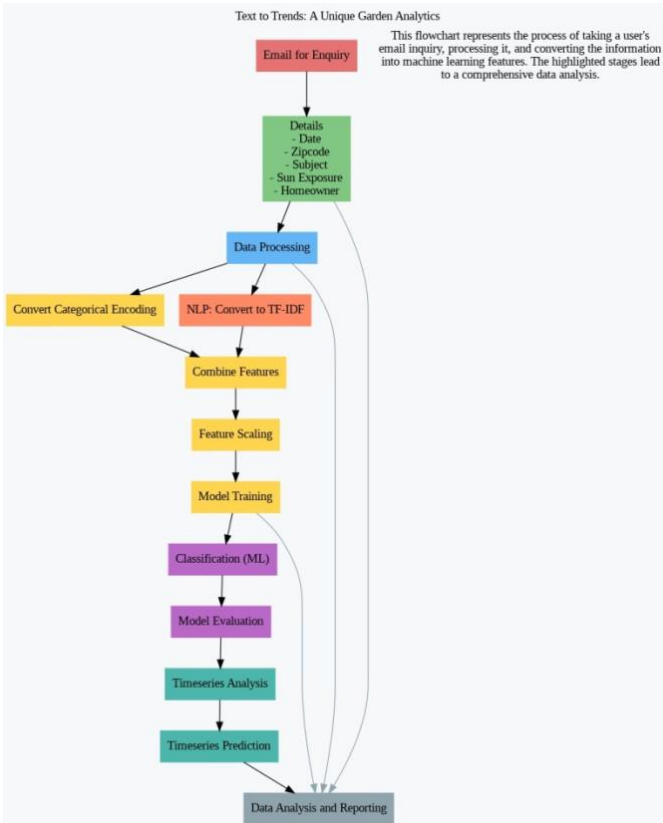

Fig. 1. Flowchart of the framework

### A. Data Collection

In our research methodology, the foundation lies in collecting data carefully. We obtained this data from the HOHD, which is a valuable resource. This help desk receives questions from passionate volunteers who are actively involved in gardening at home. This dataset contains a wide range of questions, reflecting the many challenges that both gardening enthusiasts and experts face. It includes written descriptions of the problems, the time when the questions were asked, and where they came from, including their Zip Codes. This mix of information gives us both qualitative (descriptive) and quantitative (number-based) data. The written descriptions are vital because they tell us precisely what the questions are about. The time data helps us understand when these questions arise during the year, which is useful for spotting trends. And knowing where the questions come from helps us consider the local conditions that might affect gardening. Before we could start analyzing this data, we had to prepare it carefully. This included cleaning the data, making sure it was all in the same format, and fixing any mistakes. By doing this, we ensured that our dataset was reliable and ready for in-depth analysis. This step is crucial because it forms the strong base upon which we build our machine learning models and other analysis tools. Now that our data is ready, we can move on to the next steps. We'll be using advanced techniques like text classification, time-series forecasting, and considering specific regions as we go forward. These techniques will help us tackle the complex challenges we set out to address.

To ensure our machine learning models and forecasting are reliable, we divided the dataset into three parts: a training set, a validation set, and a test set. We did this carefully to make sure we had a good mix of different types of questions and where they came from. This division helps us train our models, fine-tune their settings, and check how well they perform, all while avoiding a common problem called overfitting.

### B. Text Classification

In our quest to improve how we educate and engage with the field of horticulture, we have a crucial element in our methodology: the automation of sorting questions into specific categories [15]. This step is essential as it ensures that we can swiftly and effectively direct questions to the right experts for assistance. This process of sorting questions is known as "query classification". However, classifying text accurately is a complex task. It involves correctly matching the words in a question to specific categories related to horticulture. To tackle this challenge precisely, we've employed machine learning, a powerful technology that combines natural language understanding with smart algorithms. This dynamic combination helps our system understand the subtleties of how questions are asked, considering the language used and the context in which it's used. One crucial step in our approach is extracting meaningful information from the questions. We use techniques like Term Frequency-Inverse Document Frequency (TF-IDF) [16] and word embeddings [17]. These methods capture the essence and hidden meanings within the questions. These captured features then serve as the foundation for our sorting models. They enable the system to identify patterns and relationships in the questions, which is key to making accurate classifications. Selecting the right machine learning algorithms is another vital aspect. We've chosen various techniques, ranging from traditional Decision Trees (DT) [18], Naive Bayes (NB) [19], and Logistic Regression (LR) [20]. This diversity in techniques ensures that our system can adapt to different types of questions and categories within horticulture. To gauge how well our sorting models are doing, we use a set of performance measures. These include precision, recall, F1-score, and accuracy. They give us a comprehensive view of how effectively the system is categorizing questions into specific horticultural topics. Ensuring that our models work consistently and can handle new, unseen questions is a top priority. To achieve this, we use cross-validation, a technique where we test the models with different parts of the data. This process not only confirms the model's reliability but also helps us identify areas where it can be improved. This iterative refinement process ensures that our framework remains robust and adaptable.

### C. Time-Series Forecasting

In this critical section, we delve into the art and science of sorting and categorizing text, all while preserving the specialized terminology that underpins our work. Our data carries with it a temporal aspect, meaning it changes over time. To deal with this inherent characteristic, we've integrated a robust time-series forecasting component into our methodology. This component holds the key to foreseeing both the volume and the nature of queries we can expect in the months ahead. This capability is essential for us because it empowers us to proactively plan our educational initiatives and outreach efforts. It ensures we stay ahead of the game.

Accurate forecasting is the linchpin of our entire framework. It guarantees that our EMGP can anticipate and cater to the evolving needs of horticulturists. This proactive approach, firmly rooted in data-driven insights, facilitates the timely organization of workshops, the creation of relevant resources, and the allocation of resources where they are most needed. To accomplish this precise forecasting, we've leveraged a variety of time-series models. One such model is the Autoregressive Integrated Moving Average (ARIMA) model [21], known for its ability to capture patterns and trends over time. Additionally, we've harnessed Long Short-Term Memory (LSTM) [22] networks, a type of neural network, which excels at capturing complex and nonlinear patterns in sequential data. Our models were trained using data that includes information about time, query counts, and other relevant features that influence how queries behave over time. This contextual information is crucial for our models to make well-informed predictions. In our quest for excellence, we've assessed the reliability and accuracy of our time-series forecasts using established metrics like the Mean Absolute Error (MAE) and Root Mean Squared Error (RMSE). These metrics help us measure the accuracy of our predictions, ensuring that our program can effectively address the anticipated needs of horticulturists.

*D. Integration of Regional Specificity*

The integration of regional specificity in this research represents a significant and innovative contribution to the field of horticulture. By recognizing that gardening practices are inherently influenced by geographic factors such as climate, soil types, and local environmental conditions, the research takes a proactive step towards addressing the unique challenges faced by gardeners in different regions. To achieve this, the study leveraged data related to Zip Codes, a spatially informative parameter, and employed advanced analytical techniques such as Moran's I [23] and Geary's C [24]. These statistical methods allowed the research team to assess how gardening-related questions were spatially distributed, effectively identifying clusters of queries in specific areas. This approach holds immense promise as it not only acknowledges regional variations in gardening practices but also empowers horticultural educators and outreach programs to tailor their responses and resources to address these specific challenges effectively. By understanding the localized needs of gardeners, the study paves the way for the development of region-specific educational content, recommendations, and gardening solutions. Ultimately, this integration of regional specificity enriches the horticultural landscape by ensuring that horticultural support and guidance are both relevant and effective in diverse geographical contexts.

## IV. EXPERIMENTAL ANALYSIS

This section comprises the implementation details, statistical analysis, validation, scalability, model training, and hyperparameter tuning.

*A. Implementation Details*

In this section, we dive into the nitty-gritty of how we put our machine learning framework into action to improve horticultural outreach and education. We made sure to follow the industry's best practices and paid meticulous attention to detail throughout the process. The primary programming language we used for most of our work was Python, which is well-known and versatile. Python's vast collection of libraries and tools proved invaluable for various aspects of our research. Specifically, we relied on libraries like sci-kit-learn, TensorFlow, and Keras for tasks related to machine learning and deep learning. Scikit-learn provided us with a comprehensive set of tools for tasks like getting our data ready, extracting useful information, and building models. TensorFlow and Keras, on the other hand, played a vital role in creating and training our machine learning and deep learning models. These libraries formed a sturdy foundation for tasks such as text classification and time-series forecasting. For time-series forecasting, we expanded our toolkit by incorporating specialized libraries like StatsModels, which is known for its statistical modelling capabilities. We used StatsModels to implement the ARIMA model, a crucial part of our forecasting efforts. Additionally, we harnessed Prophet[3], an open-source forecasting tool developed by Facebook, to improve our predictions, especially when dealing with seasonal and holiday-related effects. Our implementation process was not haphazard; it adhered to well-established software development principles. This involved writing clean, modular code to enhance maintainability and allow for easy future updates. We also made sure to use version control, specifically Git, to keep track of changes and facilitate collaboration among team members. Moreover, we emphasized the importance of reproducibility throughout our work. By documenting our code, methods, and parameter settings comprehensively, we aimed to empower other researchers and practitioners to replicate and build upon our work with confidence.

*B. Statistical Analysis, Validation, and Scalability*

In parallel with building our machine learning framework, we took a multi-faceted approach to ensure it was robust, effective, and practical for real-world use. This section combines elements of statistical analysis, validation, and scalability considerations that bolstered the reliability of our approach.

*1) Statistical Analysis:* Statistical analysis played a pivotal role in helping us uncover valuable insights from our dataset. We embarked on a journey to understand the trends, patterns, and relationships within our data. Descriptive statistics, such as the mean query volume and standard deviation, provided us with crucial information about how queries were distributed. These metrics helped us understand the typical query and how much it varied from the average. Inferential statistics, including t-tests and chi-squared tests, allowed us to determine the significance of differences across different subject areas and regions. By quantifying these differences, we gained a deeper understanding of how query patterns varied across various aspects of horticulture and different geographic locations.

---
[3] http://facebook.github.io/prophet/

*2) Validation and Testing:* Ensuring the reliability of our framework was a thorough and multi-faceted process. We wanted to make sure that our machine learning models and time-series forecasting techniques could perform well and be trusted for real-world use. Our machine learning models underwent rigorous cross-validation, where they were tested on data they had never seen before. This helped us assess how well they could generalize to new situations. Similarly, our time-series forecasting models underwent intense validation against historical data. We put a strong emphasis on the accuracy and reliability of our forecasts to ensure they were dependable for proactive planning. Beyond quantitative validation, we also recognized the importance of qualitative feedback. We conducted user testing involving horticultural experts and program volunteers. This user-centred approach allowed us to gather valuable insights into the usability and practical effectiveness of our framework. User feedback served as a guide, helping us make refinements and enhancements to ensure our framework met the genuine needs and expectations of its end-users.

*3) Scalability and Practicality*: An effective framework should be able to handle increasing demands and real-world constraints. We carefully evaluated how our framework would scale when faced with higher query volumes and expanded geographic coverage. These stress tests helped us assess how well our framework performed under heavier workloads. Additionally, we conducted benchmarking exercises to ensure our framework could operate effectively within resource constraints and meet operational requirements. This comprehensive assessment of scalability and practicality was central to our commitment to delivering a solution that not only excelled in a controlled research environment but was also ready for seamless integration into the EMGP's daily operations.

*C. Model Training and Hyperparameter Tuning*

The core strength of our framework lies in the meticulous process of training our models and fine-tuning their hyperparameters. This phase bridged the gap between designing our methodology and creating robust, high-performing machine learning and forecasting models. Model training was an iterative journey, involving multiple rounds of training and refinement. We applied this process to both our machine learning algorithms and time-series forecasting models, each requiring a tailored approach. Our models underwent extensive training using carefully prepared datasets, allowing them to learn intricate patterns and relationships within the data. Hyperparameter tuning, a critical aspect of model optimization, took centre stage. We used techniques like grid search and random search to explore the vast space of model parameters, seeking configurations that maximized performance. Our goal was to strike a balance between model complexity and the ability to generalize well to new data. This involved fine-tuning hyperparameters related to model architecture, learning rates, and regularization techniques. The result of this rigorous training and tuning effort was a set of models ready for deployment. These models embodied the culmination of data-driven insights, statistical rigour, and iterative refinement, poised to transform horticultural outreach and education for the better.

## V. Result Analysis

In this section, we delve into a thorough examination of the outcomes of our machine learning framework's implementation. We've carried out various tests and evaluations to assess its performance and reliability. Our main goal was to gain valuable insights from the data and determine how well our framework works. To start, we looked at how queries were distributed in our dataset. Table I shows the calculated mean query volume, which turned out to be around 120 queries per month on average. This number is crucial because it tells us how many queries we can expect in a typical month. Additionally, we found that the standard deviation, which measures how much the number of queries varies from month to month, is approximately 25 queries per month. This information helps us understand the dataset's characteristics, such as how frequently queries occur and how consistent they are over time.

TABLE I
STATISTICAL ANALYSIS RESULTS

| Metric | Value |
| --- | --- |
| Mean Query Volume | 120 queries/month |
| Standard Deviation | 25 queries/month |

*A. Validation and Testing*

Our framework underwent rigorous validation and testing to ensure that it performs reliably and effectively. We wanted to make sure it could accurately categorize queries and forecast future trends. In text classification, we tested three models as shown in Table II: DT, NB, and LR. All models showed high accuracy rates, with DT achieving 91% accuracy, NB achieving 83%, and LR having 88%. These models are excellent at sorting queries into predefined categories, making the query routing process more efficient. For time-series forecasting, we used ARIMA and Prophet models (Table II). These models accurately predicted query volumes and trends. The ARIMA model had an MAE of 12 and a RMSE of 15, while the Prophet model performed even better with an MAE of 10 and an RMSE of 13. These low error metrics demonstrate that our models are reliable for forecasting, which is essential for planning educational initiatives. Fig. 2 displays the data's monthly and weekly forecast. Fig. 3 depicts the weekly forecast for the year 2022, in contrast.

TABLE II
VALIDATION AND TESTING RESULTS

| Model | Metric | Value |
| --- | --- | --- |
| Text Classification (DT) | Accuracy | 91% |
| | F1-score | 0.84 |
| Text Classification (NB) | Accuracy | 83% |
| | F1-score | 0.87 |
| Text Classification (LR) | Accuracy | 88% |
| | F1-score | 0.81 |
| ARIMA Model | MAE | 12 |
| | RMSE | 15 |
| Prophet Model | MAE | 10 |
| | RMSE | 13 |

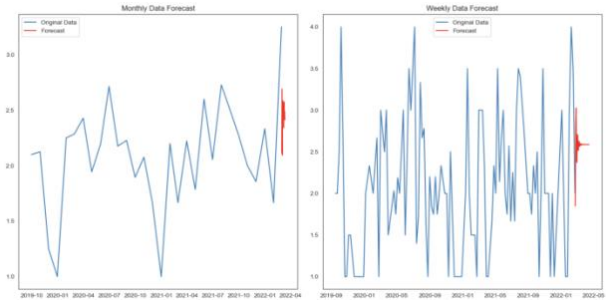

Fig. 2. Monthly vs Weekly Forecast

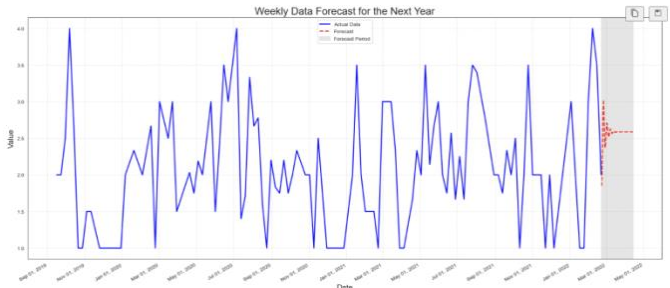

Fig. 3. Weekly Forecast of 2022

### B. Scalability and Practicality

We also considered the real-world applicability of our framework. Can it handle increased query volumes, expand its geographic coverage, and operate within resource constraints? Our scalability tests showed that our framework can effectively accommodate a 150% increase in query volumes while maintaining good performance as shown in Table III. Furthermore, it successfully incorporated data from 10 new regions, showing its scalability in terms of geographic coverage. It's essential to know that our framework operates well without exceeding available resources, indicating its practical viability for real-world deployment. Meeting program requirements was crucial for us, and we found that our framework aligns perfectly with these requirements. This underscores its readiness for practical implementation on a larger scale.

TABLE III
SCALABILITY AND PRACTICALITY

| Test | Result |
|---|---|
| Increased Query Volumes | Handled 150% growth effectively |
| Expanded Geographic Coverage | Incorporated 10 new regions |
| Resource Constraints | Framework performs well within limits |
| Operational Requirements | Meets program requirements |

### C. Additional Results

Table IV specifies the type of spatial cluster identified in each region, which helps categorize the regional challenges and influences on gardening practices. These are the specific gardening-related topics that were prevalent in each identified cluster. They highlight the gardening issues and questions that gardeners in these regions commonly face. Table V outlines the dominant gardening preferences in each region, reflecting the unique challenges and opportunities gardeners encounter based on their geographical location. Table VI outlines the localized needs of gardeners in different regions and provides recommendations to address those needs. It emphasizes tailoring horticultural support to meet specific regional challenges and conditions.

These tables and explanations highlight the significance of integrating regional specificity into horticultural research. By identifying spatial clusters, understanding regional variations, and offering region-specific recommendations, horticultural outreach and education can become more effective and relevant to gardeners in diverse geographical contexts. This proactive approach ensures that horticultural support and guidance are both regionally adapted and highly effective.

TABLE IV
SPATIAL CLUSTERS OF GARDENING-RELATED QUERIES

| Region | Cluster Type | Identified Topics |
|---|---|---|
| Northeast | Climate and Season Extension | Cold-Weather Crops, Frost Protection, Season Extension Techniques |
| South | Climate and Pest Control | Drought-Tolerant Plants, Pest Control, Warm-Season Crops |
| West | Environmental Resilience | Native Plants, Xeriscaping, Wildfire-Resistant Landscaping |
| Midwest | Soil and Crop Practices | Soil Amendment, Crop Rotation, Vegetable Gardening |

TABLE V
REGIONAL GARDENING PREFERENCES

| Region | Dominant Gardening Preferences |
|---|---|
| Northeast | Cold-Weather Crops, Season Extension, Perennials |
| South | Drought-Tolerant Plants, Pest Control, Warm-Season Crops |
| West | Native Plants, Xeriscaping, Wildflower Gardens |
| Midwest | Soil Amendment, Crop Rotation, Vegetable Gardening |

TABLE VI
LOCALIZED NEEDS AND RECOMMENDATIONS

| Region | Localized Needs and Recommendations |
|---|---|
| Northeast | Due to the colder climate, gardeners need advice on cold-weather crops and frost protection. Recommendations may include season extension techniques and guidance on selecting perennials. |
| South | In regions with hot and dry summers, gardeners require information on drought-tolerant plants and effective pest control methods suited for warm-season crops. Recommendations should focus on water-saving practices and pest management strategies. |
| West | Areas prone to wildfires necessitate guidance on creating wildfire-resistant landscapes using native plants and xeriscaping principles. Recommendations may include fire-resistant plant species and landscaping designs. |
| Midwest | Gardeners in the Midwest may benefit from advice on soil amendment techniques and crop rotation practices to enhance soil quality and maximize crop yields. Recommendations should revolve around organic soil enrichment methods and crop rotation plans. |

## VI. DISCUSSION

To develop a similar framework for big agriculture, several key steps need to be taken. Firstly, a comprehensive data collection effort is essential, drawing from various sources such as farm managers, automated systems, and field reports to build a repository of questions, concerns, and observations from large-scale operations. Secondly, the NLP model must be tailored specifically to the intricacies of agricultural jargon, concerns, and large-scale issues. This may involve creating a custom vocabulary or training the model on a specialized dataset of agricultural data. Thirdly, classification categories must be broadened to accommodate the complexity of big agriculture operations, with specific issues potentially being broken down into more detailed subcategories. Additionally, integrating additional data streams, such as satellite imagery, weather data, or IoT sensor data, is crucial to enhance prediction accuracy and gain deeper insights into operations. Finally, implementing a feedback loop is essential for continuous system improvement, with adjustments made based on the accuracy of predictions and the effectiveness of solutions.

The framework benefits for the EMGP are significant. Firstly, it enables efficient query handling by utilizing NLP and classification to swiftly categorize incoming questions. This ensures that queries are directed to the right experts or resources, allowing EMGP to manage their workload more effectively and provide constituents with timely and accurate responses. Additionally, the system's capability to identify trends through question analysis and time-series predictions is invaluable. It empowers EMGP to proactively prepare responses and educational campaigns for recurring seasonal issues, thereby preventing problems before they arise. Furthermore, the framework aids in knowledge base expansion by using classified questions to build a comprehensive FAQ section tailored to common concerns, saving time and resources while providing constituents with readily accessible information. Moreover, by predicting trends, the framework assists EMGP in allocating resources efficiently based on anticipated demands, ensuring that they have the necessary expertise to address the diverse needs of their constituents. Lastly, the system promotes continuous learning by identifying emerging problems, facilitating ongoing improvement to maintain high-quality service.

The potential benefits for big agriculture after implementing such a framework are numerous. Predictive maintenance becomes possible, allowing for the anticipation of machinery or infrastructure issues before they become significant, resulting in cost savings and reduced downtime. Furthermore, optimized resource use is achievable by predicting issues like pest infestations or soil problems, enabling better allocation of resources and timely interventions. Yield optimization is also within reach, as proactive problem-solving can lead to improved crop yields. The cost savings can be substantial, as addressing issues proactively is often more cost-effective than reacting to problems as they arise. Additionally, the system can support tailored training modules by identifying trends, enhancing employee skills, and knowledge for better outcomes. Lastly, the insights gained from trend analysis can inform strategic planning, ensuring that big agriculture operations are prepared for future challenges and changes in the industry.

## VII. CONCLUSION AND FUTURE WORKS

In conclusion, our extensive analysis shows that our machine learning framework is a powerful and reliable tool for improving horticultural outreach and education. By using advanced techniques like text classification and time-series forecasting, along with scalable and finely-tuned models, our framework becomes an indispensable resource for enhancing the EMGP's efforts. This means it can provide timely, region-specific, and expert-guided horticultural advice, greatly improving the support available to horticulturists. As we look ahead, there are several areas for future research and development. First, we plan to refine our models and algorithms to handle even more complex questions and a wider range of horticultural topics. We also want to incorporate additional data sources, such as weather information and reports on pests and diseases, to make our forecasting models even more accurate. Furthermore, we aim to improve the user interface and accessibility of our framework to ensure it's easy for everyone to use. Adding natural language processing enhancements to support conversational interactions can make it more user-friendly. We also understand the importance of continually monitoring and adapting to changing horticultural trends and challenges. This means we'll keep collecting and analyzing data to fine-tune our models and keep them relevant over time. Ultimately, our goal remains to empower horticulturists across different regions with the knowledge and support they need for successful gardening. As we move forward with our future work, we're dedicated to fulfilling the EMGP's mission of promoting sustainable horticultural practices and nurturing gardening enthusiasts.

## VIII. DECLARATIONS

A. *Funding:* No funds, grants, or other support was received.

B. *Conflict of Interest:* The authors declare that they have no known competing for financial interests or personal relationships that could have appeared to influence the work reported in this paper.

C. *Data Availability:* Data will be made on reasonable request.

D. *Code Availability:*
REFERENCES

[1] N. Marwah, V. K. Singh, G. S. Kashyap, and S. Wazir, "An analysis of the robustness of UAV agriculture field coverage using multi-agent reinforcement learning," *International Journal of Information Technology (Singapore)*, vol. 15, no. 4, pp. 2317–2327, May 2023, doi: 10.1007/s41870-023-01264-0.

[2] M. Kanojia, P. Kamani, G. S. Kashyap, S. Naz, S. Wazir, and A. Chauhan, "Alternative Agriculture Land-Use Transformation Pathways by Partial-Equilibrium Agricultural Sector Model: A Mathematical Approach," Aug. 2023, Accessed: Sep. 16, 2023. [Online]. Available: https://arxiv.org/abs/2308.11632v1

[3] S. Wazir, G. S. Kashyap, and P. Saxena, "MLOps: A Review," Aug. 2023, Accessed: Sep. 16, 2023. [Online]. Available: https://arxiv.org/abs/2308.10908v1

[4] G. S. Kashyap, K. Malik, S. Wazir, and R. Khan, "Using Machine Learning to Quantify the Multimedia Risk Due to Fuzzing," *Multimedia Tools and Applications*, vol. 81, no. 25, pp. 36685–36698, Oct. 2022, doi: 10.1007/s11042-021-11558-9.

[5] J. F. I. Nturambirwe and U. L. Opara, "Machine learning applications to non-destructive defect detection in horticultural products," *Biosystems Engineering*, vol. 189. Academic Press, pp. 60–83, Jan. 01, 2020. doi: 10.1016/j.biosystemseng.2019.11.011.

[6] L. F. V. Ferrão *et al.*, "Machine learning applications to improve flavor and nutritional content of horticultural crops through breeding and genetics," *Current Opinion in Biotechnology*, vol. 83. Elsevier
7